\documentclass[runningheads]{llncs}

\usepackage{eccv}

\usepackage{eccvabbrv}

\usepackage[pagebackref,breaklinks,colorlinks,citecolor=eccvblue]{hyperref}
\usepackage{graphicx,verbatim}
\usepackage{booktabs}
\usepackage{amssymb}
\usepackage{pifont}
\usepackage{amsmath}
\usepackage{xcolor}
\usepackage[table]{xcolor}
\usepackage{algorithm}
\usepackage{algpseudocode}
\usepackage{wrapfig}
\usepackage{float} 
\usepackage{caption}
\usepackage{enumitem}
\usepackage{orcidlink}
\usepackage{multirow}
\usepackage[dvipsnames]{xcolor}

\usepackage[accsupp]{axessibility}  % Improves PDF readability for those with disabilities.

\DeclareUnicodeCharacter{25CF}{$\bullet$}
\DeclareUnicodeCharacter{25CB}{$\circ$}

\newcommand{\blue}[1]{\textcolor{blue}{#1}{}}
\newcommand{\cyan}[1]{\textcolor{cyan}{#1}{}}

\newcommand{\st}{\scriptsize}

\begin{document}

% ---------------------------------------------------------------
% TODO REVIEW: Replace with your title
\title{Posterior Samplings are Missing Modalities Generators for Medical Image Translation} 

% TODO REVIEW: If the paper title is too long for the running head, you can set
% an abbreviated paper title here. If not, comment out.
\titlerunning{Posterior Samplings are Missing Modalities Generators}

% TODO FINAL: Replace with your author list. 
% Include the authors' ORCID for the camera-ready version, if at all possible.
\authorrunning{J. Kim}
% \author{Jonghun Kim\orcidlink{0009-0002-2790-2090}}
\author{Jonghun Kim}
% \and
% Hyunjin Park \orcidlink{0000-0001-5681-8918}}
% Department of Electrical and Computer Engineering
% \institute{Sungkyunkwan University}
\institute{Department of Electrical and Computer Engineering, \\ Sungkyunkwan University, Suwon, Korea \\
\email{iproj2@skku.edu}}
% \email{\{iproj2,hyunjinp\}@skku.edu}}

\maketitle

\begin{abstract}
Magnetic resonance imaging comes in various modality contrasts that provide complementary anatomical and pathological information. Complete multimodal acquisitions are often unavailable due to time and protocol constraints. This leads to real-world datasets with missing modalities, where conventional medical image translation methods are typically limited to fixed source-target settings or require retraining for each observed source-target pair. We propose a unified framework that formulates missing-modality generation as a linear inverse problem under a joint distribution and solves it via posterior sampling with a flow matching model. By learning a joint prior over the complete modality set, our method can reconstruct arbitrary missing modalities at inference time by guiding the sampling trajectory to enforce measurement consistency with observed modalities. We further mitigate inter-modality error propagation in multi-target generation by adopting a many-to-one sampling strategy. Experiments on BraTS and IXI datasets show that our method achieves the best performance over baselines across most missing-modality scenarios. In downstream tumor segmentation, synthesized images from our method result in higher segmentation performance, indicating better preservation of clinically relevant structures. Our code is available at \href{https://github.com/jongdory/PS-MIT}{github.com/jongdory/PS-MIT}.
\end{abstract}
\section{Introduction}
\label{sec:intro}

Magnetic resonance imaging (MRI) is an essential non-invasive tool in clinical workflows \cite{katti2011magnetic}. 
MRI comes in multiple modalities, such as T1-weighted, T2-weighted, and T2-weighted fluid-attenuated inversion recovery (FLAIR) images, that provide complementary anatomical and pathological information \cite{menze2014multimodal}. 
For example, T1-weighted images emphasize anatomy and tissue boundaries, while T2-weighted and FLAIR images provide contrast for identifying pathological changes, such as edema or lesions, by highlighting differences in water content \cite{filippi2016mri, de1992mr}. Analyzing these multimodal images is necessary for a comprehensive understanding of a patient's condition. 
Despite their clinical utility, acquiring a complete set of modalities is impractical due to time and protocol constraints. 
For instance, FLAIR requires long repetition and inversion times to suppress cerebrospinal fluid signals \cite{menze2014multimodal, kuijf2019standardized}. 
As a result, real-world datasets often contain incomplete data where certain modalities are missing.
 
To address the missing-modality problem, medical image translation has been studied with generative models, evolving from generative adversarial network (GAN)-based methods to diffusion-based methods \cite{yu2019ea, NEURIPS2021_0f281810, dalmaz2022resvit, kim2024adaptive, ozbey2023unsupervised, arslan2025self}. 
However, existing work focuses on fixed translation settings, translating a commonly available source modality into a frequently missing target modality. 
Such fixed formulations do not reflect the diverse missing-modality scenarios encountered in practice and often require training separate models for different input-output pairs. 
In fixed settings, if a specific required modality is missing, models cannot perform synthesis and cannot benefit from additional available modalities that potentially improve generation quality. 
Consequently, these methods are inflexible and inefficient to deploy across tasks.
An ideal framework should support synthesis from arbitrary observed modalities within a unified model and dynamically fuse information across available modalities \cite{sharma2019missing, xiong2025learning, dorent2025unified}.

\begin{figure}[t]
    \centering
    \includegraphics[width=0.99\textwidth]{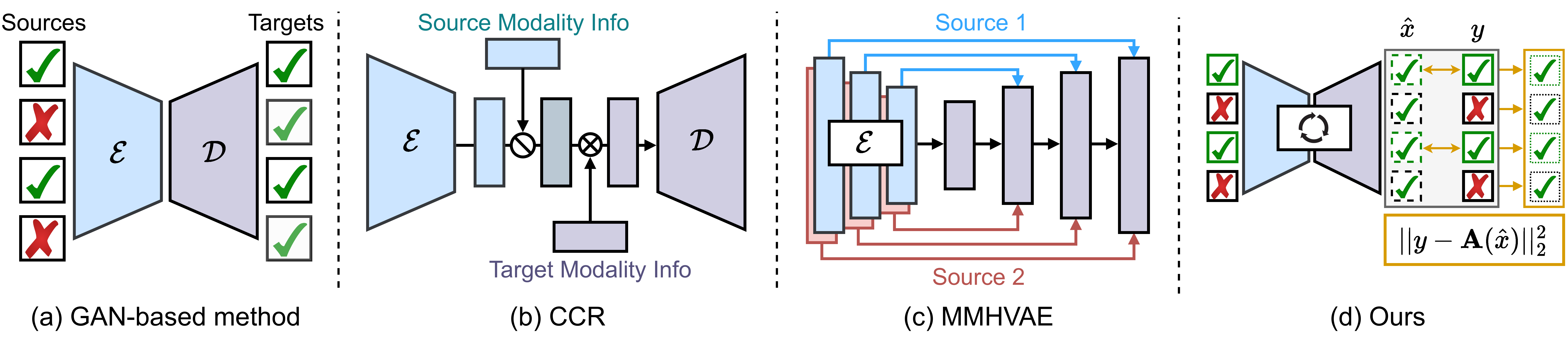}
    \caption{Overview of representative missing-modality generation frameworks.
            \textbf{(a) GAN-based method} \cite{sharma2019missing} concatenates observed modalities channel-wise and handles missing channels with zeros.
            \textbf{(b) CCR} \cite{xiong2025learning} removes source information from encoder features and injects target information.
            \textbf{(c) MMHVAE} \cite{dorent2025unified} generates targets from latent variables using features from modality-specific encoders.
            \textbf{(d) Ours} uses a flow-based model and posterior sampling. The sampling trajectory is guided by the current estimate $\hat{\boldsymbol{x}}$ and measurements $\boldsymbol{y}$ to enforce consistency with observed modalities.}
    \label{fig1}
    \vspace{-6pt}
\end{figure}

To address these requirements, unified approaches such as MM-GAN \cite{sharma2019missing}, ResViT \cite{dalmaz2022resvit}, CCR \cite{xiong2025learning}, and MMHVAE \cite{dorent2025unified} have been proposed to generate missing modalities from partially observed inputs (Fig~\ref{fig1}).
For instance, MM-GAN \cite{sharma2019missing} and ResViT \cite{dalmaz2022resvit} rely on zero imputation to handle incomplete data. 
CCR \cite{xiong2025learning} attempts to explicitly disentangle contrast and content representations, but in medical imaging, these two components are highly related. MRI contrast depends on the intrinsic relaxation times of specific anatomical structures. Thus, decoupling them could result in the loss of critical pathological features, such as tumor edema \cite{brown2014magnetic}.
Forcing this separation risks losing inherent correlations causing structural distortions in the synthesized images.
MMHVAE \cite{dorent2025unified} utilizes a multi-level hierarchical structure to generate high-resolution images, which requires a complex design to compress and reconstruct information at each layer managing dependencies between latent variables. 
Furthermore, the optimization of multiple sub-networks and the hierarchical structure makes it challenging to balance the various loss terms.

Unlike previous methods that depend on complex architectural or explicit disentanglement, we treat missing modality generation as a linear inverse problem under a joint distribution and solve it via posterior sampling \cite{chung2023diffusion, wang2023zeroshot} with an iterative generative model \cite{song2021scorebased, ho2020denoising, liu2023flow, lipman2023flow}. 
We train a flow-based model \cite{lipman2023flow} to learn the joint prior over the complete modality set. 
At inference time, we start from noise and guide the sampling trajectory using the observed modalities to reconstruct missing modalities. 
Our method has the advantage of addressing all missing scenarios by learning a standard joint prior. 
By learning a single joint prior, our method supports diverse observed-missing combinations and leverages all available modalities.

Our contributions are as follows:
\begin{itemize}[noitemsep, topsep=0pt, partopsep=0pt, parsep=0pt]
\item[◦] We propose a novel approach for missing modality generation by training a joint generative model and employing posterior sampling.
\item[◦] We identify the error accumulation when generating multiple modalities simultaneously and propose a many-to-one sampling strategy to mitigate this.
\item[◦] Our approach successfully generates missing modalities and outperforms baseline models on the BraTS and IXI datasets.
\end{itemize}

\section{Related Works}
\label{sec:related}

\noindent \textbf{Medical Image Translation.}
Medical image translation aims to synthesize missing modalities. Prior work has adopted GANs \cite{NIPS2014_5ca3e9b1} and, more recently, diffusion models \cite{ho2020denoising}.
Early GAN-based approaches primarily relied on the Pix2Pix \cite{isola2017image}, with notable advancements such as Ea-GAN \cite{yu2019ea} for edge guidance, ResViT \cite{dalmaz2022resvit} for transformer-based architectures, and RegGAN \cite{NEURIPS2021_0f281810} for addressing spatial misalignments. More recently, diffusion-based models have emerged, including ALDM \cite{kim2024adaptive}, which utilizes style information within the latent space. SynDiff \cite{ozbey2023unsupervised} integrates the diffusion process with adversarial training. SelfRDB \cite{arslan2025self} employs a diffusion-bridge mechanism. However, these models are typically constrained to fixed one-to-one or many-to-one settings.
To support diverse missing-modality scenarios (Fig.~\ref{fig1}), unified frameworks such as MM-GAN \cite{sharma2019missing}, CCR \cite{xiong2025learning}, and MMHVAE \cite{dorent2025unified} have been proposed. They broaden the translation setting beyond fixed pairs but often introduce additional architectural or optimization complexity.
In contrast, our approach avoids specialized architectures by learning a joint generative prior over the complete multimodal set and reconstructing missing modalities via posterior sampling at inference time.

\vspace{3pt}
\noindent \textbf{Inverse Problems.} 
Inverse problems aim to recover a true signal $\boldsymbol{x}$ from a measurement $\boldsymbol{y}$ corrupted by a forward model. 
This forward model usually represents a physical, computational, or statistical process that maps the true signal to the measurement space. This process often introduces noise, distortion, or information loss.
Recent methods leverage pre-trained diffusion and flow-based models as generative priors $p(\boldsymbol{x})$. By integrating the forward model and measurement, these approaches sample from the posterior $p(\boldsymbol{x}|\boldsymbol{y})$, often utilizing the log-likelihood gradient $\nabla_{\boldsymbol{x}} \log p(\boldsymbol{y}|\boldsymbol{x})$, to solve tasks like inpainting and super-resolution without task-specific retraining \cite{daras2024survey}. 
Early works focused on diffusion models that operate in pixel space \cite{chung2022improving, chung2023diffusion, wang2023zeroshot, chung2024decomposed}. 
These methods were later extended to latent space to take advantage of pre-trained latent models \cite{rout2023solving, chung2024prompttuning, zhang2025improving}. 
More recently, flow-based models have been proposed for solving linear inverse problems \cite{Patel_2025_ICCV, Kim_2025_ICCV}.
DPS \cite{chung2023diffusion} guides the reverse diffusion process using an approximation of the likelihood gradient to improve measurement consistency. DDNM \cite{wang2023zeroshot} enforces data consistency via a projection-based decomposition into range and null spaces. PSLD \cite{rout2023solving} applies gradient-based guidance in the latent space to enable efficient high-resolution restoration.
These posterior-sampling methods have also been extended to flow models. FlowChef \cite{Patel_2025_ICCV} introduces a guidance term into the reverse ordinary differential equation (ODE) to enforce data consistency, while FlowDPS \cite{Kim_2025_ICCV} generalizes posterior sampling to affine flows by decomposing Euler steps using a generalized Tweedie's formula \cite{efron2011tweedie}.
These methods can be categorized by their constraint mechanisms. DPS, FlowChef, and FlowDPS use gradient-based soft constraints. In contrast, DDNM uses a projection-based hard constraint to enforce measurement consistency. PSLD uses a hybrid approach that combines gradient-based and projection-based constraints. While previous studies have primarily focused on improving the perceptual quality in natural image domain, our goal is to improve image fidelity in the medical domain. We compare the performance of various posterior sampling methods to identify the optimal method.
% Here, we adapt these posterior-sampling ideas to generate missing modalities from a pre-trained flow-based model and compare against representative baselines.
\section{Preliminaries}

\subsection{Conditional Flow Matching} 
Conditional flow matching (CFM) learns a probability path $p_t$ that transports a noise distribution $p_0$ to a data distribution $q$ via a straight-line trajectory. We define $\boldsymbol{x}_t$ as:
\begin{equation}
\boldsymbol{x}_t = (1 - t)\boldsymbol{x}_0 + t\boldsymbol{x}_1,
\end{equation}
where $t \in [0, 1]$, $\boldsymbol{x}_0 \sim p_0$ is noise and $\boldsymbol{x}_1 \sim q$ is a target sample. The corresponding conditional vector field is $u_t(\boldsymbol{x}_1 |\boldsymbol{x}_0)=\boldsymbol{x}_1-\boldsymbol{x}_0$. 
Our model takes a multi-channel image $\boldsymbol{x}_t$, where each channel corresponds to a different modality. The training objective is:
\begin{equation}
\mathcal{L}_\text{CFM}(\theta) = \mathbb{E}_{t, q(\boldsymbol{x}_1), p_0(\boldsymbol{x}_0)} \| u_\theta(\boldsymbol{x}_t, t) - (\boldsymbol{x}_1 - \boldsymbol{x}_0) \|^2, 
\end{equation}
where $\theta$ denotes the learnable parameters of the neural network for $u_\theta$ and $t$ is sampled uniformly from $\mathcal{U}(0, 1)$. This objective trains the vector field that transforms a Gaussian distribution into a multimodal data distribution.

\subsection{Problem Formulation}
The task of reconstructing missing modalities can be mathematically formulated as a linear inverse problem. A typical linear inverse problem aims to recover an unknown clean sample $\boldsymbol{x}_1 \in \mathbb{R}^d$ from a set of measurements $\boldsymbol{y}$:
\begin{equation}
\boldsymbol{y} = \mathcal{A}\boldsymbol{x}_1 + \boldsymbol{n}\in \mathbb{R}^m,
\label{eq3}
\end{equation}
where $\mathcal{A} \in \mathbb{R}^{m \times d}$ is the measurement operator and $\boldsymbol{n} \sim \mathcal{N}(0, \sigma^2 \mathbf{I})$ denotes Gaussian noise. Inverse problems are inherently ill-posed. There are multiple solutions $\boldsymbol{x}$ that satisfy Eq. \ref{eq3}. We represent a multimodal sample as a multi-channel image $\boldsymbol{x}\in\mathbb{R}^{C\times H\times W}$, where $C$ denotes the number of modalities. The masking operator $\mathcal{A}$ selects the observed input modality channels.
In our work, $\mathcal{A}$ is a channel-selection operator determined by the available modalities at inference time and $\boldsymbol{y}$ contains only observed channels. 
\section{Methods}
\label{sec:method}

\begin{figure}[t]
    \centering
    \includegraphics[width=0.99\textwidth]{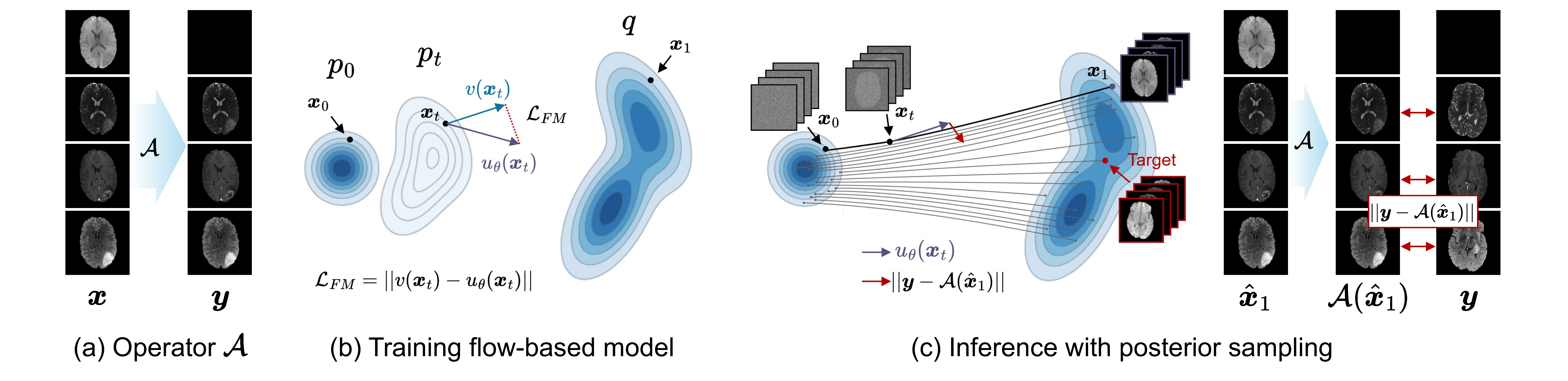}
    \caption{Overview of the proposed training and inference pipeline.
            \textbf{(a)} The measurement operator $\mathcal{A}$ masks missing modality channels and $\boldsymbol{y}$ denotes the observed modalities.
            \textbf{(b)} The flow-based model learns a time-dependent vector field that transports noise to data.
            \textbf{(c)} At inference, missing modalities are reconstructed via posterior sampling guided to minimize the difference between the projected estimate $\mathcal{A}(\hat{\boldsymbol{x}}_1)$ and $\boldsymbol{y}$. }
    \label{fig2}
\end{figure}

We formulate missing-modality generation as a linear inverse problem and solve it using a pre-trained flow-based model. 
Specifically, we learn a joint generative model over multi-modal images by representing all modalities as channels of a single multi-channel sample. 
At inference time, missing modalities are reconstructed via posterior sampling (Fig.~\ref{fig2}). 
The model learns a probability path from noise to data by predicting the corresponding vector field at each time step $t$. 
During sampling, we enforce measurement consistency using the residual between the projected estimate $\mathcal{A}(\hat{\boldsymbol{x}}_1)$ and the measurement $\boldsymbol{y}$, guiding the trajectory to synthesize missing channels while preserving observed modalities.

\subsection{Measurement-Consistent Sampling}
In medical image translation, fidelity is the main priority and takes precedence over perceptual quality or diversity \cite{rassmann2025regression}. 
Any structural inconsistency or hallucination in the synthesized output can severely compromise its clinical utility. 
To mitigate these risks and ensure anatomical accuracy, we employ a sampling strategy that enforces strict data consistency. 
Flow-based models generate samples by solving the probability flow ODE. The sampling update is defined as:
\begin{equation}
d\boldsymbol{x}_t = v_\theta(\boldsymbol{x}_t, t)\, dt, \qquad \boldsymbol{x}_{t+dt} = \boldsymbol{x}_t + v_\theta(\boldsymbol{x}_t,t)\,dt,
\end{equation}
where $v_\theta$ denotes the vector field predicted by the model. 
To reconstruct missing modalities while preserving the observed modalities, we introduce two interventions during sampling. 
First, we anchor the observed channels by replacing the predicted vector field in those channels with a measurement-induced vector field. 
Following inpainting-style strategies, we enforce the ideal velocity toward the ground truth in the observed channels, while letting the unobserved (missing) channels evolve according to the model prediction \cite{lugmayr2022repaint, martin2025pnpflow}.
Let $C$ be the number of modality channels and let $\boldsymbol{y}$ denote the observed measurements.
We use a binary channel mask $\boldsymbol{m} \in \{0,1\}^C$ to indicate which channels are observed or missing.
The modified vector field $\hat{v}_t$ used for the update step is defined as:
\begin{equation}
\label{eq5}
\hat{v}_t = \boldsymbol{m} \odot (\boldsymbol{y} - \boldsymbol{x}_0) + (1 - \boldsymbol{m}) \odot u_\theta(\boldsymbol{x}_t, t).   
\end{equation}
This keeps observed channels consistent, while allowing the model to synthesize missing channels. Second, we apply a posterior-sampling guidance term minimizing a $\|\boldsymbol{y} - \mathcal{A}(\hat{\boldsymbol{x}}_{1})\|_2^2$, which steers the trajectory toward the measurement-consistent set:
\begin{equation}    
\label{eq6}
\boldsymbol{x}_t \leftarrow \boldsymbol{x}_t - s\nabla_{\hat{\boldsymbol{x}}_{1}}\|\boldsymbol{y} - \mathcal{A}(\hat{\boldsymbol{x}}_{1})\|_2^2,
\end{equation}
where $\nabla_{\hat{\boldsymbol{x}}_{1}}$ is the gradient of $\hat{\boldsymbol{x}}_{1}$ and $s$ denotes the step size.

\begin{figure}[t]
    \centering
    \includegraphics[width=0.99\textwidth]{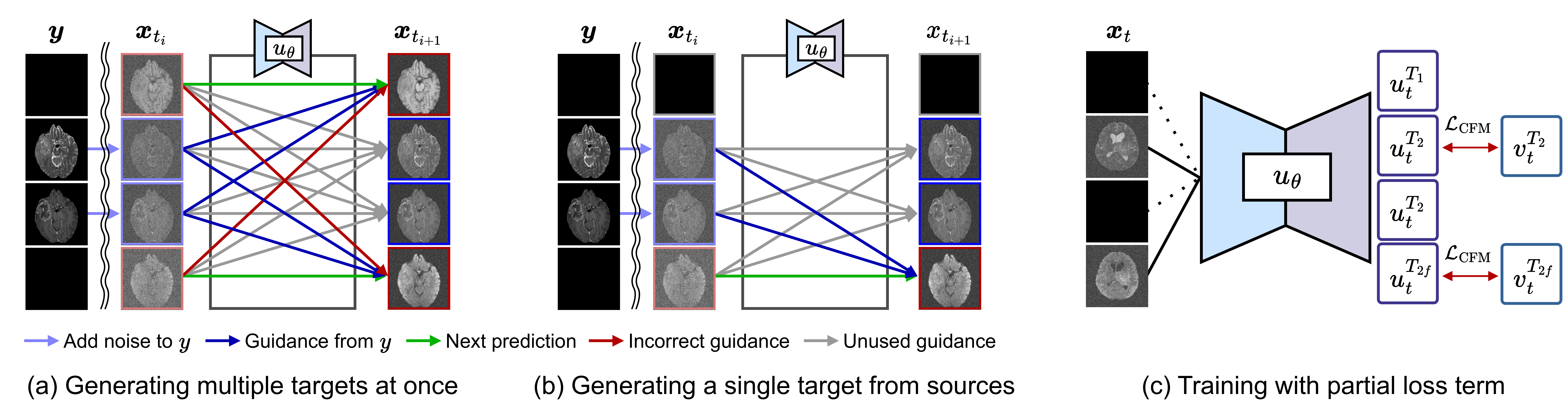}
    \caption{Mitigating inter-modality error propagation.
            \textbf{(a)} Jointly synthesizing multiple modalities may propagate errors across channels (red arrows).
            \textbf{(b)} Many-to-one sampling generates one target modality at a time, avoiding interference from other synthesized channels.
            \textbf{(c)} For training, our model uses random combinations of observed-target scenarios and calculates the vector field of the observed channels. }
    \label{fig3}
\end{figure}

\subsection{Mitigating Inter-Modality Error Propagation}
The proposed method can simultaneously synthesize all modalities regardless of the number of missing modalities.
However, in scenarios with multiple missing modalities, particularly in many-to-many generation tasks, error accumulation may occur.
During generation, deviations from the target trajectory in a given modality can propagate to other modalities, potentially affecting the generation paths of other target modalities. 
The generation process should rely solely on real, observed information. 
However, the joint generation process allows irrelevant information or errors originating from an intermediate stage of unverified modalities to propagate to other target modalities (Fig.~\ref{fig3}(a)).
These accumulated errors can lead to severe structural inconsistencies.
In contrast, as shown in Fig.~\ref{fig3}(b), the single-target generation process fully utilizes the observed modality information without incorrect guidance. 

\vspace{3pt}
\noindent \textbf{Training Strategy.}
The training process must be flexible to handle diverse target generation scenarios. 
The model should use only the observed modalities as inputs for each target translation task.
For example, in T2$\rightarrow$FLAIR scenario (Fig.~\ref{fig3}(c)), only the T2 and FLAIR channels remain active while other channels are masked.
During the sampling process, FLAIR is generated with the observed modality T2.
To handle all scenarios, the model is trained on the vector field of the input modalities.
Missing modalities are zeroed to be excluded from the computation, as zero-filling an input channel is equivalent to its removal \cite{zeiler2014visualizing, goodfellow2016deep}. The convolution operation relies on the multiplication of inputs and kernels, thus setting an input channel to zero is equivalent to pruning its corresponding weight channel.
We scale the output based on the number of inputs after the convolution step to prevent variability due to varying numbers of channels.
During training, we employ a stochastic masking strategy where up to two input channels are randomly masked. We define a scenario mask $\boldsymbol{m}_s \in \{0, 1\}^C$ to indicate which channels are used. The model is optimized using a partial loss as shown in Fig.~\ref{fig3}.

\vspace{3pt}
\noindent \textbf{Sampling Strategy.} For many-

\begin{wrapfigure}{r}{0.56\columnwidth}
    \vspace{-4.7\baselineskip}
    \begin{minipage}{0.56\columnwidth}
        \begin{algorithm}[H]
            \captionsetup{font=footnotesize}
            \caption{Many-to-one sampling}
            \label{algo1}
            \begin{algorithmic}[1]
                \State \textbf{Input:} Flow-based model $u_\theta$, input noise sample $\boldsymbol{x}_0 \sim N(0, I)$, mask $\boldsymbol{m}$, scenario mask $\boldsymbol{m}_s$, measurements $\boldsymbol{y}$, and step size $s$.
                \State \textbf{for} $t \in \{0...1\}$ \textbf{do}
                \begingroup
                    \State \hspace{\algorithmicindent} $\boldsymbol{x}_t \leftarrow \boldsymbol{m}_s\odot\boldsymbol{x}_t$ \qquad // scenario masking
                    \State \hspace{\algorithmicindent} $v \leftarrow u_\theta(\boldsymbol{x}_t, t)$ 
                    \State \hspace{\algorithmicindent} $dt \leftarrow 1/T$ 
                    \State \hspace{\algorithmicindent} $\hat{\boldsymbol{x}}_1 \leftarrow \boldsymbol{x}_t +(1 - t)\cdot v$  
                    \State \hspace{\algorithmicindent} $\boldsymbol{x}_t \leftarrow \boldsymbol{x}_t - s\nabla_{\hat{\boldsymbol{x}}_{1}}\|\boldsymbol{y} - \mathcal{A}(\hat{\boldsymbol{x}}_{1})\|_2^2$  \ // Eq. \ref{eq6}
                    \State \hspace{\algorithmicindent} $v^{obs} \leftarrow \boldsymbol{y} - \boldsymbol{x}_0$ \quad // observed vector field
                    \State \hspace{\algorithmicindent} $v \leftarrow \boldsymbol{m}\odot v^{obs} + (1 - \boldsymbol{m})\odot v$ \ \ // Eq. \ref{eq5}
                    \State \hspace{\algorithmicindent} $\hat{\boldsymbol{x}}_{t+1} \leftarrow \boldsymbol{x}_t + dt \cdot v$ 
                \endgroup
                \State \Return $\boldsymbol{x}_1$
            \end{algorithmic}
        \end{algorithm}
    \end{minipage}
    \vspace{-2\baselineskip}
\end{wrapfigure}

\noindent to-one sampling, we ensure that only the observed and target modalities influence the generation process with scenario mask $\boldsymbol{m}_s$.
For example, if there are two observed modalities and one target modality, all other channels are zeroed so that only these three channels are active. We perform posterior sampling guided by the observed modalities and overwrite the observed vector fields using the mask $\boldsymbol{m}$. The many-to-one sampling procedure is summarized in Algorithm~\ref{algo1}.
Notably, in the absence of  $\boldsymbol{m}_s$, the procedure becomes many-to-many sampling.

\vspace{3pt}
\noindent \textbf{Expectation Approximation Sampling.}
In physical MRI acquisition, noise is inevitable \cite{gudbjartsson1995rician}. Generative models learn this inherent noise during training. In practice, repeatedly acquiring MRIs and averaging the signals can effectively suppress this noise \cite{roemer1990nmr, aja2013review}. Ideally, as the number of acquired samples approaches infinity, the average converges to a noise-free image. Similarly, since flow-based models sample non-deterministically from random noise, each generated sample contains different random noise. Leveraging this, previous research \cite{rassmann2025regression} proposed Expectation-Approximation (ExpA) sampling, which adapts the physical principle of signal averaging to generative models to suppress noise. The noise-free image $\boldsymbol{x}'$ can be approximated by averaging multiple generated samples with the number of ExpA sampling $N_\text{Ex}$:
\begin{equation} 
    \mathbb{E}_{\hat{\boldsymbol{x}} \sim p_\theta} = \underset{N_\text{Ex} \rightarrow \infty} \lim \hat{\bar{\boldsymbol{x}}}_{N_\text{Ex}}
    \approx \underset{N_\text{Ex} \rightarrow \infty} \lim \bar{\boldsymbol{x}}_{ N_\text{Ex}} = \boldsymbol{x}',
\end{equation}
where $p_\theta$ is the probability distribution based on networks, $\hat{\boldsymbol{x}}$ is a generated image, $\hat{\bar{\boldsymbol{x}}}$ is the average of generated images, and $\bar{\boldsymbol{x}}$ is the average of measured images.

\section{Experiments}

\subsection{Experimental Setup}

\noindent \textbf{Datasets.}
We used the BraTS2023 \cite{menze2014multimodal,bakas2017advancing,baid2021rsna} and IXI \footnote{https://brain-development.org/ixi-dataset/} datasets for model training and validation.
From the BraTS, we used T1w, T2w, T1ce, and FLAIR images. A total of 1140 subjects were randomly selected, with 1000 used for training, 40 for validation, and 100 for testing. All images were skull-stripped and resampled to 1 mm isotropic resolution.
From the IXI dataset, we used T1w, T2w, and proton density (PD) images. We randomly selected 200 subjects, allocating 120 for training, 20 for validation, and 60 for testing. All images were preprocessed for skull stripping, affine registration, and resampling to a spacing of $0.9375 \times 0.9375 \times 1.2 \text{mm}^3$.

\vspace{3pt}
\noindent \textbf{Implementation Details.}
All experiments were performed using PyTorch \cite{paszke2019pytorch} in Python with CUDA 12.1 and the Adam \cite{kingma2014adam} optimizer was employed for training. A single NVIDIA A100 80GB GPU was used. For the sampling, the number of function evaluations (NFE) was set to 40, while $N_\text{Ex}$ was set to 10. We adopted PSLD \cite{rout2023solving} as the posterior sampling method.

\vspace{3pt}
\noindent \textbf{Model Architectures.} The backbone network architecture was a UNet \cite{ronneberger2015u} with both the encoder and decoder comprising four ResNet \cite{he2016deep} blocks using 128, 256, 512, and 512 channels. 

\vspace{3pt}
\noindent \textbf{Metrics.} To evaluate image fidelity, PSNR and SSIM were utilized. The perceptual quality was assessed using FID \cite{heusel2017gans} and LPIPS \cite{zhang2018unreasonable} to quantify the perception-distortion trade-off \cite{blau2018perception}.

\vspace{3pt}
\noindent \textbf{Comparison Methods.}
We compared the unified models, MM-GAN \cite{sharma2019missing}, ResViT \cite{dalmaz2022resvit}, CCR \cite{xiong2025learning}, and MMHVAE \cite{dorent2025unified} as baseline models for generating missing modalities. 
Additionally, we compared various posterior sampling methods of DPS \cite{chung2023diffusion}, DDNM \cite{wang2023zeroshot}, PSLD \cite{rout2023solving}, FlowChef \cite{Patel_2025_ICCV}, and FlowDPS \cite{Kim_2025_ICCV}. 
The supplement contains details of the posterior sampling methods. 

\subsection{Results with Comparison Methods}
\noindent \textbf{Results on the BraTS.}  We evaluated our method across all possible missing modality scenarios on the BraTS, with quantitative results summarized in Table \ref{table1}. 
In most cases, the proposed method outperformed the baselines. 
Performance improved incrementally as the number of source (observed) modalities increased from one to three.
For single-source scenarios, T1 synthesis was most effective when using T1ce or T2 as the source. 
For other targets, the best source modalities were FLAIR for T2 synthesis, T1 for T1ce synthesis, and T2 for FLAIR synthesis. 
Similarly, in two-source scenarios, the highest performance was achieved by the combinations of \{T2, T1ce\} for T1, \{T1, FLAIR\} for T2, and \{T1, T2\} for both T1ce and FLAIR. 
Using T1ce or FLAIR as the source did not always result in the highest scores, even though they are rich in tumor information. 
This is because the metrics are calculated on the whole image, rather than just tumor regions. This suggests that other structures and contrast outside the tumor are more heavily weighted when using other modalities as the source. This tendency was also observed in the baseline models.

\begin{table} [t]
    \setlength{\tabcolsep}{5pt} 
    \centering
    \caption{Quantitative performance of MRI synthesis with baselines on the BraTS. 
    $\text{T}_1$, $\text{T}_2$, $\text{T}_{1ce}$, and $\text{T}_{2f}$ denote T1-weighted, T2-weighted, T1 contrast-enhanced, and FLAIR, respectively.
    Sources denote the observed modalities. Mean and standard deviation values are presented. 
    \blue{Blue} indicates the best performance. 
    \underline{Underline} indicates statistically significant differences compared to all others ($p < 0.05$). }
    \vspace{-3pt}
    \label{table1}
    \centering
    \scalebox{0.52}{
        \begin{tabular}{ c | c @{\hspace{2.5pt}} c @{\hspace{2.5pt}} c @{\hspace{2pt}} c|ccccc|ccccc}
            \toprule
             \multirow{2}{*}{\rotatebox{0}{\textbf{Target}}} & \multicolumn{4}{c}{\textbf{Sources}}  & \multicolumn{5}{|c}{\textbf{PSNR (dB)} $\uparrow$}  & \multicolumn{5}{|c}{\textbf{SSIM} (\%) $\uparrow$}  \\
            \cmidrule(lr){2-5} \cmidrule(lr){6-10} \cmidrule(lr){11-15} 
             & $\text{T}_1$ \ & $\text{T}_2$ \ & $\text{T}_{1c}$ & $\text{T}_{2f}$ & MMGAN & ResViT & CCR & MMHVAE & \textbf{Ours} & MMGAN & ResViT & CCR & MMHVAE & \textbf{Ours} \\
            \midrule
            \rowcolor{gray!10} $\text{T}_1$ & ○ & ● & ○ & ○ & 24.36 \st(2.63) & 23.05 \st(2.12) & 24.02 \st(2.51) & 22.67 \st(2.49) & \blue{\underline{25.56 \st(2.39)}} & 90.28 \st(3.24) & 89.61 \st(3.37) & 89.92 \st(3.20) & 88.10 \st(3.47) &            \blue{90.52 \st(3.12)}  \\
                                            & ○ & ○ & ● & ○ & 25.24 \st(3.09) & 23.86 \st(2.83) & 25.15 \st(2.63) & 23.92 \st(2.60) & \blue{\underline{25.56 \st(2.14)}} & 87.70 \st(3.27) & 87.40 \st(3.31) & 87.59 \st(3.01) & 89.15 \st(3.57) &            \blue{88.17 \st(2.84)}  \\
            \rowcolor{gray!10}              & ○ & ○ & ○ & ● & 23.96 \st(2.24) & 22.95 \st(1.60) & 23.63 \st(2.37) & 21.49 \st(2.51) & \blue{\underline{24.66 \st(2.61)}} & 88.02 \st(2.85) & 88.01 \st(2.83) & 87.74 \st(3.02) & 84.28 \st(3.09) &            \blue{88.27 \st(3.06)}  \\
                                            & ○ & ● & ● & ○ & 26.01 \st(3.26) & 24.01 \st(3.14) & 25.50 \st(2.83) & 24.61 \st(2.85) & \blue{\underline{27.15 \st(2.43)}} & 91.94 \st(3.20) & 90.82 \st(3.52) & 91.73 \st(2.86) & 90.63 \st(3.52) &             \blue{91.96 \st(2.80)} \\
            \rowcolor{gray!10}              & ○ & ● & ○ & ● & 24.90 \st(2.55) & 23.83 \st(2.34) & 24.47 \st(2.58) & 23.05 \st(2.74) & \blue{\underline{26.86 \st(2.33)}} & 90.93 \st(2.98) & 89.16 \st(3.13) & 90.49 \st(2.89) & 89.02 \st(2.99) & \blue{\underline{92.03 \st(2.76)}} \\
                                            & ○ & ○ & ● & ● & 26.09 \st(3.11) & 24.10 \st(2.89) & 25.35 \st(2.67) & 24.21 \st(2.54) & \blue{\underline{27.33 \st(2.25)}} & 91.29 \st(3.07) & 90.40 \st(3.10) & 91.07 \st(2.71) & 89.98 \st(3.21) & \blue{\underline{91.66 \st(2.70)}} \\
            \rowcolor{gray!10}              & ○ & ● & ● & ● & 26.23 \st(3.22) & 24.66 \st(3.06) & 25.54 \st(2.87) & 24.76 \st(2.73) & \blue{\underline{27.84 \st(2.24)}} & 92.15 \st(3.10) & 91.25 \st(3.42) & 91.80 \st(2.74) & 90.35 \st(2.99) & \blue{\underline{92.95 \st(2.71)}} \\
            \midrule
                               $\text{T}_2$ & ● & ○ & ○ & ○ & 23.20 \st(2.35) & 23.28 \st(1.96) & 22.33 \st(2.54) & 20.53 \st(2.39) & \blue{\underline{23.56 \st(2.70)}} & 88.31 \st(6.92) & 88.57 \st(3.45) & 87.76 \st(7.06) & 85.10 \st(7.22) &            \blue{88.47 \st(3.89)}  \\
            \rowcolor{gray!10}              & ○ & ○ & ● & ○ & 22.82 \st(2.26) & 22.72 \st(1.75) & 22.40 \st(2.18) & 20.84 \st(2.14) & \blue{\underline{23.09 \st(1.89)}} & 87.44 \st(6.55) & 88.06 \st(2.71) & 86.69 \st(6.48) & 83.88 \st(6.56) &            \blue{85.46 \st(2.83)}  \\
                                            & ○ & ○ & ○ & ● & 23.16 \st(2.05) & 23.24 \st(2.01) & 22.05 \st(2.04) & 20.88 \st(1.84) & \blue{\underline{23.71 \st(2.23)}} & 87.06 \st(6.34) & 87.69 \st(2.54) & 86.70 \st(6.66) & 83.66 \st(6.56) & \blue{\underline{88.15 \st(3.02)}} \\
            \rowcolor{gray!10}              & ● & ○ & ● & ○ & 23.95 \st(2.53) & 23.61 \st(1.84) & 23.11 \st(2.35) & 21.25 \st(2.34) & \blue{\underline{25.27 \st(2.30)}} & 89.16 \st(6.72) & 89.38 \st(3.03) & 88.53 \st(6.70) & 85.76 \st(6.93) & \blue{\underline{90.27 \st(3.38)}} \\
                                            & ● & ○ & ○ & ● & 24.45 \st(2.42) & 24.42 \st(2.18) & 23.53 \st(2.38) & 21.81 \st(2.26) & \blue{\underline{26.22 \st(2.42)}} & 90.07 \st(6.78) & 90.52 \st(3.15) & 89.74 \st(6.88) & 87.04 \st(6.85) & \blue{\underline{91.87 \st(3.41)}} \\
            \rowcolor{gray!10}              & ○ & ○ & ● & ● & 24.46 \st(2.44) & 24.31 \st(1.98) & 23.53 \st(2.24) & 22.13 \st(2.24) & \blue{\underline{25.52 \st(1.98)}} & 89.59 \st(6.54) & 90.07 \st(2.55) & 89.10 \st(6.58) & 86.49 \st(6.58) & \blue{\underline{90.89 \st(2.97)}} \\
                                            & ● & ○ & ● & ● & 24.89 \st(2.67) & 24.49 \st(2.08) & 24.00 \st(2.39) & 22.32 \st(2.38) & \blue{\underline{26.06 \st(2.18)}} & 90.67 \st(6.75) & 90.92 \st(2.91) & 90.21 \st(6.74) & 87.51 \st(6.81) & \blue{\underline{92.60 \st(3.16)}} \\
            \midrule
        \rowcolor{gray!10} $\text{T}_{1ce}$ & ● & ○ & ○ & ○ & 22.70 \st(2.09) & 22.88 \st(2.04) & 21.92 \st(1.96) & 22.11 \st(2.59) & \blue{\underline{24.15 \st(3.28)}} & 87.15 \st(3.71) & 85.62 \st(3.36) & 86.98 \st(3.71) & 85.74 \st(3.81) &            \blue{87.28 \st(3.70)}  \\
                                            & ○ & ● & ○ & ○ & 22.34 \st(2.19) & 21.46 \st(1.79) & 21.75 \st(1.89) & 20.98 \st(2.03) & \blue{\underline{22.87 \st(2.08)}} & 86.23 \st(2.92) & 83.69 \st(2.62) & 85.25 \st(2.81) & 83.99 \st(2.86) &            \blue{86.60 \st(2.96)}  \\
            \rowcolor{gray!10}              & ○ & ○ & ○ & ● & 22.12 \st(1.88) & 21.48 \st(1.72) & 21.58 \st(1.66) & 20.68 \st(2.01) & \blue{\underline{22.72 \st(2.54)}} & 84.85 \st(3.03) & 83.26 \st(2.83) & 83.95 \st(3.11) & 80.71 \st(3.05) &            \blue{85.00 \st(3.18)}  \\
                                            & ● & ● & ○ & ○ & 23.32 \st(2.47) & 22.84 \st(2.12) & 22.42 \st(1.88) & 22.53 \st(2.53) & \blue{\underline{24.95 \st(3.10)}} & 88.46 \st(3.26) & 85.77 \st(3.00) & 87.40 \st(3.40) & 86.54 \st(3.47) & \blue{\underline{89.38 \st(3.34)}} \\
            \rowcolor{gray!10}              & ● & ○ & ○ & ● & 22.99 \st(2.36) & 22.78 \st(2.05) & 22.18 \st(1.79) & 22.54 \st(2.50) & \blue{\underline{24.62 \st(3.14)}} & 87.98 \st(3.46) & 85.25 \st(3.25) & 87.27 \st(3.48) & 85.72 \st(3.92) & \blue{\underline{89.17 \st(3.26)}} \\
                                            & ○ & ● & ○ & ● & 22.69 \st(2.13) & 22.15 \st(1.88) & 22.11 \st(1.81) & 21.40 \st(2.06) & \blue{\underline{23.91 \st(2.61)}} & 87.23 \st(2.85) & 85.65 \st(2.66) & 86.22 \st(2.74) & 84.07 \st(2.85) & \blue{\underline{88.04 \st(2.89)}} \\
            \rowcolor{gray!10}              & ● & ● & ○ & ● & 23.30 \st(2.64) & 22.95 \st(2.21) & 22.49 \st(1.85) & 22.76 \st(2.18) & \blue{\underline{25.00 \st(3.12)}} & 88.69 \st(3.21) & 86.14 \st(3.03) & 87.82 \st(3.18) & 86.06 \st(3.41) & \blue{\underline{90.15 \st(3.08)}} \\
            \midrule
                            $\text{T}_{2f}$ & ● & ○ & ○ & ○ & 20.73 \st(2.27) & 20.30 \st(2.01) & 20.16 \st(1.93) & 19.73 \st(1.49) & \blue{\underline{21.45 \st(2.22)}} & 84.26 \st(3.08) & 84.32 \st(2.77) & 83.32 \st(3.15) & 82.16 \st(3.29) &            \blue{84.70 \st(3.42)}  \\
            \rowcolor{gray!10}              & ○ & ● & ○ & ○ & 22.63 \st(2.67) & 22.48 \st(2.30) & 21.39 \st(2.43) & 20.96 \st(2.04) & \blue{\underline{22.98 \st(2.32)}} & 85.34 \st(3.33) & 85.42 \st(2.86) & 85.50 \st(3.47) & 84.28 \st(3.56) &            \blue{85.55 \st(3.13)}  \\
                                            & ○ & ○ & ● & ○ & 21.73 \st(2.33) & 22.41 \st(2.14) & 20.64 \st(1.96) & 20.20 \st(1.74) & \blue{\underline{22.27 \st(1.95)}} & 83.49 \st(3.22) & 83.62 \st(2.86) & 83.07 \st(3.31) & 81.53 \st(3.08) &            \blue{84.02 \st(2.97)}  \\
            \rowcolor{gray!10}              & ● & ● & ○ & ○ & 23.22 \st(2.74) & 23.91 \st(2.54) & 21.81 \st(2.56) & 21.86 \st(2.11) & \blue{\underline{24.51 \st(2.33)}} & 87.81 \st(3.08) & 87.73 \st(2.86) & 87.01 \st(3.14) & 85.77 \st(3.06) & \blue{\underline{88.46 \st(3.24)}} \\
                                            & ● & ○ & ● & ○ & 22.37 \st(2.35) & 22.76 \st(2.12) & 21.04 \st(2.16) & 20.41 \st(1.70) & \blue{\underline{23.18 \st(2.05)}} & 85.78 \st(3.01) & 85.76 \st(2.74) & 84.52 \st(3.20) & 82.80 \st(3.04) & \blue{\underline{86.79 \st(3.23)}} \\
            \rowcolor{gray!10}              & ○ & ● & ● & ○ & 23.27 \st(2.51) & 23.92 \st(2.35) & 22.00 \st(2.51) & 21.92 \st(2.02) & \blue{\underline{24.29 \st(2.71)}} & 87.68 \st(3.01) & 87.55 \st(2.81) & 86.72 \st(3.21) & 85.13 \st(3.03) & \blue{\underline{88.09 \st(3.17)}} \\
                                            & ● & ● & ● & ○ & 23.55 \st(2.63) & 23.93 \st(2.32) & 22.10 \st(2.63) & 21.75 \st(1.85) & \blue{\underline{24.72 \st(2.70)}} & 88.27 \st(2.99) & 87.88 \st(2.75) & 87.29 \st(3.12) & 84.57 \st(2.83) & \blue{\underline{89.11 \st(3.18)}} \\
            \bottomrule
        \end{tabular}
    }
\end{table}

Statistical analysis via t-tests revealed that our method achieved significant improvements over others in terms of PSNR across all tasks ($p < 0.05$). However, significant differences in SSIM were primarily observed in two-source tasks, with less improvement in single-source scenarios. To investigate this, we visualized results for three specific cases (T1 $\rightarrow$ T2, \{T1, T2\} $\rightarrow$ FLAIR, and \{T1, T2, FLAIR\} $\rightarrow$ T1ce), as shown in Fig. \ref{fig4}. In the single-source case, while our method produced noise-free images, they appeared slightly blurrier compared to the baselines, which likely affected the SSIM scores. Nevertheless, our model successfully reconstructed images similar to the ground truth without structural hallucinations.
As the number of source modalities increased, the synthesized images became sharper. Unlike competing models that failed to accurately represent tumor regions, our method showed the highest fidelity to the ground truth in terms of overall contrast, structural detail. Specifically, T1ce synthesis, the most challenging task due to the difficulty of predicting enhancing tumor and tumor core regions, highlighted the robustness of our approach. While baselines failed to capture tumor details, often underestimating enhancing regions (MMGAN, CCR, MMHVAE) or producing distortions (ResViT), our model achieved the most accurate and realistic depictions of the tumor.

\begin{figure}[t]
    \begin{minipage} {\textwidth}
        \centering
        \includegraphics[width=0.99\textwidth]{figs/fig4.pdf}
        \caption{Qualitative comparison of our method with baselines for synthesizing in various scenarios.
        \textbf{Top}: T1 $\rightarrow$ T2 synthesis. The bounding boxes show zoomed-in views of the brain tumor and ventricular regions.
        \textbf{Middle}: T1, T2 $\rightarrow$ FLAIR synthesis. The bounding boxes show zoomed-in views of the brain tumor and ventricular regions.
        \textbf{Bottom}: T1, T2, FLAIR $\rightarrow$ T1ce synthesis. The bounding boxes show zoomed-in views of the brain tumor region.
        }
        \label{fig4}
    \end{minipage}
    % \vspace{3pt}
    \begin{minipage} {\textwidth}
        \setlength{\tabcolsep}{5.5pt} 
        \centering
        \captionof{table}{Quantitative performance of MRI synthesis with baselines on the IXI. Mean and standard deviation values are presented. \blue{Blue} indicates the best performance. \underline{Underline} indicates statistical significance compared to all others ($p < 0.05$).}
        \vspace{6pt}
        \label{table2}
        \centering
        \scalebox{0.53}{
            \begin{tabular}{c|c @{\hspace{1pt}} c @{\hspace{1pt}}c|ccccc|ccccc}
                \toprule
                \multirow{2}{*}{\rotatebox{0}{\textbf{Target}}} & \multicolumn{3}{c}{\textbf{Sources}}  & \multicolumn{5}{|c}{\textbf{PSNR (dB)} $\uparrow$}  & \multicolumn{5}{|c}{\textbf{SSIM} (\%) $\uparrow$}  \\
                \cmidrule(lr){2-4} \cmidrule(lr){5-9} \cmidrule(lr){10-14} 
                 & $\text{T}_1$ \ & $\text{T}_2$ \ & $\text{PD}$ & MMGAN & ResViT & CCR & MMHVAE & \textbf{Ours} & MMGAN & ResViT & CCR & MMHVAE & \textbf{Ours} \\
                \midrule
                \rowcolor{gray!10} $\text{T}_1$ & ○ & ● & ○ & 23.84 \st(2.48) & 24.00 \st(2.71) & 21.64 \st(1.64) & 23.76 \st(2.63) & \blue{\underline{24.64 \st(1.53)}} & 90.10 \st(4.98) & 90.29 \st(4.81) & 89.84 \st(4.61) & 88.28 \st(4.72) & \blue{\underline{90.76 \st(2.70)}} \\
                                                & ○ & ○ & ● & 23.31 \st(2.22) & 23.89 \st(2.57) & 22.17 \st(1.66) & 22.50 \st(2.14) &           \blue{{23.94 \st(1.09)}} & 89.13 \st(4.55) & 89.50 \st(4.93) & 86.49 \st(3.73) & 84.38 \st(4.14) & \blue{\underline{90.33 \st(1.46)}} \\
                \rowcolor{gray!10}              & ○ & ● & ● & 24.24 \st(2.65) & 24.79 \st(2.94) & 21.68 \st(1.80) & 24.09 \st(2.68) & \blue{\underline{25.93 \st(1.96)}} & 91.16 \st(5.08) & 91.33 \st(5.38) & 89.97 \st(4.77) & 87.77 \st(4.66) & \blue{\underline{91.72 \st(3.18)}} \\
                \midrule
                 $\text{T}_2$                   & ● & ○ & ○ & 21.20 \st(1.69) & 20.82 \st(1.42) & 20.11 \st(1.21) & 21.86 \st(1.78) & \blue{\underline{22.78 \st(1.21)}} & 88.45 \st(4.21) & 88.79 \st(4.36) & 86.69 \st(3.86) & 88.28 \st(4.31) & \blue{\underline{89.51 \st(2.14)}} \\
                \rowcolor{gray!10}              & ○ & ○ & ● & 21.45 \st(1.39) & 22.92 \st(1.44) & 20.51 \st(1.06) & 23.35 \st(1.43) &           \blue{{23.49 \st(0.99)}} & 89.99 \st(2.79) & 89.63 \st(3.05) & 88.09 \st(2.92) & 90.11 \st(2.97) &           \blue{{90.18 \st(1.40)}} \\
                                                & ● & ○ & ● & 22.18 \st(1.62) & 23.70 \st(1.55) & 22.27 \st(1.60) & 24.24 \st(2.04) & \blue{\underline{25.38 \st(2.02)}} & 91.16 \st(3.01) & 91.22 \st(3.26) & 91.36 \st(3.35) & 91.58 \st(3.14) &           \blue{{91.80 \st(3.12)}} \\
                \midrule
                \rowcolor{gray!10} $\text{PD}$  & ● & ○ & ○ & 23.19 \st(1.84) & 23.57 \st(1.90) & 23.11 \st(1.57) & 23.24 \st(1.78) & \blue{\underline{24.85 \st(1.10)}} & 87.79 \st(3.93) & 88.03 \st(4.16) & 86.52 \st(3.65) & 86.67 \st(3.64) & \blue{\underline{88.16 \st(2.48)}} \\
                                                & ○ & ● & ○ & 26.07 \st(1.62) & 26.39 \st(1.95) & 25.66 \st(2.24) & 26.03 \st(1.66) & \blue{\underline{26.64 \st(1.20)}} & 91.04 \st(2.92) & 90.61 \st(3.03) & 89.47 \st(2.79) & 90.63 \st(2.90) & \blue{\underline{91.64 \st(1.95)}} \\
                \rowcolor{gray!10}              & ● & ● & ○ & 26.31 \st(1.60) & 26.49 \st(2.18) & 26.76 \st(2.23) & 26.13 \st(1.93) & \blue{\underline{27.72 \st(1.69)}} & 92.35 \st(2.94) & 92.39 \st(3.30) & 90.60 \st(2.96) & 90.81 \st(2.91) &           \blue{{92.48 \st(2.59)}} \\
                \bottomrule
            \end{tabular}
        }
    \end{minipage}
\end{figure}

\vspace{3pt}
\noindent \textbf{Results on the IXI.} We compared the missing modality synthesis results for all possible scenarios on the IXI, with the quantitative results summarized in Table \ref{table2}. Unlike on the BraTS, which consists of tumor patient images, IXI contains normal brain scans. Consistent with the trends observed on the BraTS, both the baselines and the proposed method achieved higher performance when two source modalities were provided compared to a single-source scenario. In single-source tasks, the most effective source modalities were identified as T2 for T1 synthesis, PD for T2 synthesis, and T2 for PD synthesis. These results suggest that the synergy of multiple observed modalities significantly enhances the fidelity of the generated sequences, even in healthy brain structures. Our model also showed the highest performance in missing modality generation on IXI, proving the robustness of our method.

\begin{table} [t]
    \setlength{\tabcolsep}{5pt} 
    \centering
    \caption{Quantitative performance of MRI synthesis with different posterior sampling methods on the BraTS. Mean and standard deviation values are presented. \blue{Blue} indicates the best performance. \cyan{Cyan} indicates the second best performance.}
    \label{table3}
    \centering
    \scalebox{0.515}{
        \begin{tabular}{ c | c @{\hspace{2.5pt}} c @{\hspace{2.5pt}} c @{\hspace{2pt}} c|ccccc|ccccc}
            \toprule
            \multirow{2}{*}{\rotatebox{0}{\textbf{Target}}} & \multicolumn{4}{c}{\textbf{Sources}}  & \multicolumn{5}{|c}{\textbf{PSNR (dB)} $\uparrow$}  & \multicolumn{5}{|c}{\textbf{SSIM} (\%) $\uparrow$}  \\
            \cmidrule(lr){2-5} \cmidrule(lr){6-10} \cmidrule(lr){11-15} 
             & $\text{T}_1$ \ & $\text{T}_2$ \ & $\text{T}_{1c}$ & $\text{T}_{2f}$ &  DPS & DDNM & PSLD & FlowChef & FlowDPS & DPS & DDNM & PSLD & FlowChef & FlowDPS  \\
            \midrule
            \rowcolor{gray!10} $\text{T}_1$ & ○ & ● & ○ & ○ & 24.21 \st(2.01) & \cyan{24.26 \st(1.93)} & \blue{25.56 \st(2.39)} & 24.21 \st(1.99) & 24.10 \st(1.89) & 87.31 \st(2.64) & \cyan{87.37 \st(2.62)} & \blue{90.52 \st(3.12)} & 87.29 \st(2.65) & 85.49 \st(2.97) \\
                                            & ○ & ○ & ● & ○ & 25.32 \st(1.88) & 25.32 \st(1.88) & \blue{25.56 \st(2.14)} & 25.31 \st(1.88) & \cyan{25.43 \st(1.92)} & 88.34 \st(2.45) & \blue{88.36 \st(2.46)} & 88.17 \st(2.84) & \cyan{88.35 \st(2.42)} & 88.16 \st(2.51) \\
            \rowcolor{gray!10}              & ○ & ○ & ○ & ● & 23.91 \st(2.39) & 23.96 \st(2.37) & \blue{24.66 \st(2.61)} & 23.94 \st(2.35) & \cyan{24.04 \st(2.21)} & 86.56 \st(2.70) & \cyan{86.61 \st(2.76)} & \blue{88.27 \st(3.06)} & 86.59 \st(2.75) & 84.52 \st(6.29) \\
                                            & ○ & ● & ● & ○ & 27.01 \st(2.07) & 26.99 \st(2.07) & \blue{27.15 \st(2.43)} & 26.94 \st(2.08) & \cyan{27.11 \st(2.10)} & 91.86 \st(2.71) & 91.83 \st(2.69) & \blue{91.96 \st(2.80)} & 91.85 \st(2.69) & \cyan{91.90 \st(2.70)} \\
            \rowcolor{gray!10}              & ○ & ● & ○ & ● & 26.17 \st(2.37) & 26.12 \st(2.36) & \blue{26.86 \st(2.33)} & 26.02 \st(2.31) & \cyan{26.32 \st(2.18)} & \cyan{91.35 \st(2.66)} & 91.32 \st(2.68) & \blue{92.03 \st(2.76)} & 91.30 \st(2.69) & 90.54 \st(2.70) \\
                                            & ○ & ○ & ● & ● & 26.68 \st(1.97) & 26.71 \st(1.97) & \blue{27.33 \st(2.25)} & 26.73 \st(1.97) & \cyan{26.84 \st(2.00)} & 91.05 \st(2.59) & 91.08 \st(2.54) & \blue{91.66 \st(2.70)} & \cyan{91.10 \st(2.60)} & 90.75 \st(5.16) \\
            \rowcolor{gray!10}              & ○ & ● & ● & ● & 27.47 \st(2.11) & 27.47 \st(2.08) & \blue{27.84 \st(2.24)} & 27.47 \st(2.07) & \cyan{27.65 \st(2.05)} & 92.72 \st(2.67) & 92.73 \st(2.64) & \blue{92.95 \st(2.71)} & 92.73 \st(2.65) & \cyan{92.74 \st(2.68)} \\
            \midrule
                               $\text{T}_2$ & ● & ○ & ○ & ○ & 22.15 \st(2.36) & \cyan{22.16 \st(2.36)} & \blue{23.56 \st(2.70)} & 22.15 \st(2.41) & 22.11 \st(2.27) & 86.06 \st(3.39) & \cyan{86.09 \st(3.41)} & \blue{88.47 \st(3.89)} & 86.08 \st(3.44) & 85.85 \st(3.40) \\
            \rowcolor{gray!10}              & ○ & ○ & ● & ○ & 21.45 \st(1.83) & \cyan{21.58 \st(1.80)} & \blue{23.09 \st(1.89)} & 21.50 \st(1.87) & 21.49 \st(1.82) & 83.73 \st(2.53) & \cyan{83.83 \st(2.59)} & \blue{85.46 \st(2.83)} & 83.76 \st(2.65) & 83.62 \st(2.63) \\
                                            & ○ & ○ & ○ & ● & 23.02 \st(2.07) & 22.96 \st(2.14) & \blue{23.71 \st(2.23)} & 23.00 \st(2.07) & \cyan{23.06 \st(1.97)} & \cyan{86.08 \st(2.82)} & 85.94 \st(2.88) & \blue{88.15 \st(3.02)} & 86.02 \st(2.77) & 85.87 \st(2.83) \\
            \rowcolor{gray!10}              & ● & ○ & ● & ○ & 23.65 \st(2.10) & 23.68 \st(2.03) & \blue{25.27 \st(2.30)} & \cyan{23.72 \st(2.03)} & 23.70 \st(2.06) & 88.40 \st(3.25) & \cyan{88.46 \st(3.25)} & \blue{90.27 \st(3.38)} & 88.43 \st(3.27) & 88.38 \st(3.30) \\
                                            & ● & ○ & ○ & ● & 25.34 \st(2.09) & 25.36 \st(2.08) & \blue{26.22 \st(2.42)} & 25.36 \st(2.07) & \cyan{25.56 \st(2.24)} & 90.78 \st(3.27) & 90.80 \st(3.27) & \blue{91.87 \st(3.41)} & \cyan{90.82 \st(3.27)} & 90.75 \st(3.33) \\
            \rowcolor{gray!10}              & ○ & ○ & ● & ● & 24.62 \st(1.77) & 24.72 \st(1.66) & \blue{25.52 \st(1.98)} & 24.60 \st(1.78) & \cyan{24.92 \st(1.71)} & 89.21 \st(2.80) & \cyan{89.26 \st(2.78)} & \blue{90.89 \st(2.97)} & 89.21 \st(2.79) & 89.21 \st(2.77) \\
                                            & ● & ○ & ● & ● & 25.68 \st(1.93) & 25.67 \st(1.87) & \cyan{26.06 \st(2.18)} & 25.68 \st(1.90) & \blue{26.11 \st(1.87)} & 91.80 \st(3.11) & 91.80 \st(3.09) & \blue{92.60 \st(3.16)} & 91.78 \st(3.12) & \cyan{91.82 \st(3.11)} \\
            \midrule
        \rowcolor{gray!10} $\text{T}_{1ce}$ & ● & ○ & ○ & ○ & 23.68 \st(2.98) & 23.70 \st(2.95) & \blue{24.15 \st(3.28)} & 23.70 \st(2.95) & \cyan{23.74 \st(2.51)} & 86.90 \st(3.15) & 86.95 \st(3.19) & \blue{87.28 \st(3.70)} & \cyan{86.97 \st(3.19)} & 86.77 \st(3.18) \\
                                            & ○ & ● & ○ & ○ & 22.10 \st(2.30) & \cyan{22.14 \st(2.24)} & \blue{22.87 \st(2.08)} & 22.09 \st(2.24) & 22.03 \st(1.69) & 84.89 \st(2.70) & \cyan{84.95 \st(2.66)} & \blue{86.60 \st(2.96)} & 84.88 \st(2.63) & 81.18 \st(3.30) \\
            \rowcolor{gray!10}              & ○ & ○ & ○ & ● & 22.29 \st(2.42) & 22.32 \st(2.40) & \blue{22.72 \st(2.54)} & 22.30 \st(2.39) & \cyan{22.39 \st(1.91)} & 84.22 \st(2.84) & \cyan{84.23 \st(2.84)} & \blue{85.00 \st(3.18)} & 84.21 \st(2.87) & 83.11 \st(2.97) \\
                                            & ● & ● & ○ & ○ & \cyan{24.62 \st(2.87)} & 24.60 \st(2.85) & \blue{24.95 \st(3.10)} & 24.60 \st(2.83) & 24.50 \st(2.83) & \cyan{88.98 \st(3.13)} & 88.97 \st(3.12) & \blue{89.38 \st(3.34)} & 88.97 \st(3.11) & 88.93 \st(3.05) \\
            \rowcolor{gray!10}              & ● & ○ & ○ & ● & 24.18 \st(2.81) & \cyan{24.21 \st(2.82)} & \blue{24.62 \st(3.14)} & 24.18 \st(2.80) & 24.15 \st(2.79) & 88.48 \st(2.99) & \cyan{88.54 \st(2.97)} & \blue{89.17 \st(3.26)} & 88.49 \st(2.98) & 88.46 \st(2.98) \\
                                            & ○ & ● & ○ & ● & 23.42 \st(2.48) & 23.44 \st(2.53) & \blue{23.91 \st(2.61)} & 23.44 \st(2.52) & \cyan{23.51 \st(2.44)} & 87.57 \st(2.74) & 87.60 \st(2.78) & \blue{88.04 \st(2.89)} & 87.60 \st(2.78) & \cyan{87.70 \st(2.70)} \\
            \rowcolor{gray!10}              & ● & ● & ○ & ● & 24.77 \st(2.82) & 24.83 \st(2.83) & \blue{25.00 \st(3.12)} & 24.76 \st(2.82) & \cyan{24.89 \st(2.74)} & \cyan{89.85 \st(2.97)} & 89.83 \st(2.97) & \blue{90.15 \st(3.08)} & 89.83 \st(2.97) & 89.83 \st(2.98) \\
            \midrule
                            $\text{T}_{2f}$ & ● & ○ & ○ & ○ & 20.29 \st(2.05) & \cyan{20.32 \st(2.06)} & \blue{21.45 \st(2.22)} & 20.24 \st(2.02) & 20.25 \st(2.02) & 83.77 \st(3.13) & 83.75 \st(3.10) & \blue{84.70 \st(3.42)} & \cyan{83.78 \st(3.15)} & 83.69 \st(3.06) \\
            \rowcolor{gray!10}              & ○ & ● & ○ & ○ & 21.54 \st(2.05) & \cyan{21.57 \st(2.11)} & \blue{22.98 \st(2.32)} & 21.48 \st(2.09) & 21.15 \st(2.14) & 83.53 \st(2.87) & \cyan{83.60 \st(2.91)} & \blue{85.55 \st(3.13)} & 83.48 \st(2.86) & 82.86 \st(2.69) \\
                                            & ○ & ○ & ● & ○ & 21.24 \st(1.97) & \cyan{21.26 \st(1.89)} & \blue{22.27 \st(1.95)} & 21.23 \st(1.93) & 21.12 \st(1.87) & 83.77 \st(2.82) & 83.87 \st(2.87) & \blue{84.02 \st(2.97)} & 83.81 \st(2.86) & \cyan{83.88 \st(2.85)} \\
            \rowcolor{gray!10}              & ● & ● & ○ & ○ & 23.85 \st(2.68) & \cyan{23.86 \st(2.58)} & \blue{24.51 \st(2.33)} & 23.85 \st(2.63) & 23.82 \st(2.67) & 88.01 \st(3.18) & 87.98 \st(3.16) & \blue{88.46 \st(3.24)} & 87.95 \st(3.19) & \cyan{88.05 \st(3.15)} \\
                                            & ● & ○ & ● & ○ & 22.06 \st(1.90) & \cyan{22.12 \st(1.86)} & \blue{23.18 \st(2.05)} & 22.05 \st(1.94) & 22.07 \st(1.90) & 86.09 \st(3.06) & \cyan{86.14 \st(3.07)} & \blue{86.79 \st(3.23)} & 86.09 \st(3.12) & 86.12 \st(3.12) \\
            \rowcolor{gray!10}              & ○ & ● & ● & ○ & 23.92 \st(2.14) & 23.88 \st(2.28) & \blue{24.29 \st(2.71)} & \cyan{23.95 \st(2.18)} & 23.93 \st(2.24) & 87.75 \st(3.07) & 87.75 \st(3.04) & \blue{88.09 \st(3.17)} & 87.78 \st(3.06) & \cyan{87.81 \st(3.05)} \\
                                            & ● & ● & ● & ○ & 24.27 \st(2.45) & 24.22 \st(2.36) & \blue{24.72 \st(2.70)} & 24.34 \st(2.37) & \cyan{24.50 \st(2.40)} & 88.88 \st(3.13) & 88.87 \st(3.07) & \blue{89.11 \st(3.18)} & 88.87 \st(3.09) & \cyan{88.98 \st(3.05)} \\
            \bottomrule
        \end{tabular}
    }
\end{table}

\subsection{Results with Posterior Sampling methods}  
We evaluated various posterior sampling methods on the BraTS dataset, with the quantitative results presented in Table \ref{table3}. Depending on the selected posterior sampling methods, Eq. \ref{eq6} in Algorithm \ref{algo1} is adapted accordingly (details in the supplement).
Across most tasks, PSLD consistently achieved the highest performance. DPS, FlowChef, and FlowDPS employ gradient-based soft constraints; among these, FlowDPS yielded the best results, ranking second overall behind PSLD. DDNM, which utilizes a projection-based hard constraint to enforce measurement consistency, was next to FlowDPS in performance.
PSLD demonstrated the highest image fidelity across most tasks by leveraging a hybrid approach that integrates both gradient-based and projection-based constraints. Specifically, PSLD achieved the best performance by computing the gradient and projection distance between the estimation and the measurement. This mechanism ensures that the synthesized images strictly adhere to the observed source modalities while effectively preserving the generative prior to seamlessly reconstruct the missing modalities.

\begin{figure}[t]
    \centering
    \includegraphics[width=0.99\textwidth]{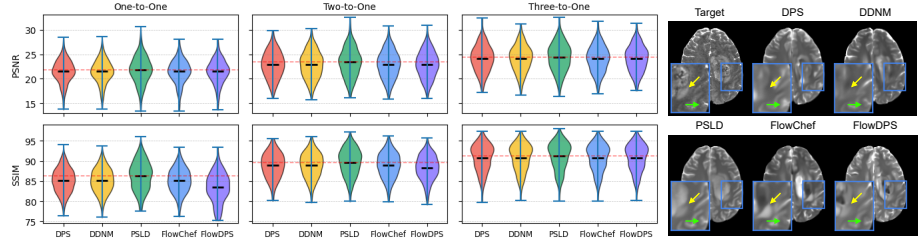}
    \caption{Illustration of synthesis results on BraTS using various posterior sampling methods.
            Left: Quantitative performance plot relative to the number of source modalities. 
            The red line shows the mean performance of PSLD, which achieved the best performance.
            Right: Qualitative comparison of T1 $\rightarrow$ T2 synthesis results with $N_\text{EX}=1$.
            The yellow arrow indicates edema, while the green arrow indicates the cortex region.}
    \label{fig5}
\end{figure}

We also analyzed the performance of each method relative to the number of available source modalities, as illustrated in Fig. \ref{fig5}-left. PSLD consistently outperformed the other methods across one, two, three-source scenarios. 
The SSIM gap between PSLD and the baselines was most apparent when fewer source modalities were available, but it gradually diminished as the number of sources increased. 
The PSLD outperformed other methods in PSNR, though the differences were slight. Since PSNR is more affected by noise and contrast than structural metrics like SSIM, most models produced comparable results. 
This trend suggests that the synthesis process relies heavily on posterior sampling guidance when source information is scarce, whereas it depends more on vector field replacement (Eq. \ref{eq5}) when ample source modalities are provided. 
Fig. \ref{fig5}-right presents the qualitative results of $T_1 \rightarrow T_2$ synthesis for each method. The gradient-based methods (DPS, FlowChef, and FlowDPS) struggled to generate the tumor edema region and failed to accurately depict the cortex. Although DDNM successfully reconstructed the cortex, it exhibited weak contrast in the edema region. Conversely, PSLD reconstructed both regions with the highest fidelity to the ground truth. Based on these findings, we adopted PSLD as the default posterior sampling method for our framework.

\begin{figure}[t]
    \centering
    \includegraphics[width=0.99\textwidth]{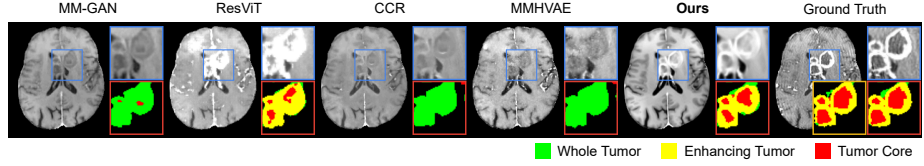}
    \caption{Illustration of tumor segmentation results with synthesized T1ce (T1, T2, FLAIR $\rightarrow$ T1ce). The blue bounding boxes shows the tumor region.The red bounding boxes shows segmentation predictionm and the yellow bounding boxes shows the ground truth.}
    \label{fig6}
\end{figure}

\subsection{Impact on Tumor Segmentation} 
\begin{wrapfigure}{r}{0.57\textwidth}
    \vspace{-2.5\baselineskip}
    \setlength{\tabcolsep}{5pt} 
    \centering
    \captionof{table}{Tumor segmentation performance with synthesized T1ce (T1, T2, FLAIR $\rightarrow$ T1ce). }
    \vspace{3pt}
    \label{table4}
    \centering
    \scalebox{0.615}{
        \begin{tabular}{l|cccc}
            \toprule
             \multicolumn{1}{c|}{\multirow{2}{*}{\textbf{Methods}}} & \multicolumn{4}{c}{\textbf{Dice (\%) $\uparrow$}} \\
              \cmidrule(lr){2-5} 
              & \textbf{WT} & \textbf{ET} & \textbf{TC} & \textbf{Avg.} \\
            \midrule
            MMGAN \cite{sharma2019missing}  & 90.75 \st(4.99) & 77.04 \st(15.14) & 62.30 \st(21.55) & 76.69 \st(10.98) \\ 
            ResViT \cite{dalmaz2022resvit}  & 91.39 \st(5.78) & 76.25 \st(16.18) & 52.01 \st(24.26) & 73.21 \st(12.65) \\ 
            CCR \cite{xiong2025learning}    & 85.75 \st(9.22) & 53.44 \st(23.03) & 38.11 \st(19.18) & 59.10 \st(16.92) \\ 
            MMHVAE \cite{dorent2025unified} & 84.25 \st(9.10) & 50.81 \st(21.51) & 36.99 \st(18.22) & 57.35 \st(16.07) \\ 
            \textbf{Ours}                   & \blue{91.53 \st(6.17)} & \blue{79.97 \st(11.32)} & \blue{63.79 \st(18.19)} & \blue{78.43 \st(9.30)} \ \\ 
            \midrule
            \rowcolor{gray!8} Ground Truth & 93.78 \st(4.20) & 94.44 \st(5.34) \ & 86.38 \st(17.98) & 91.53 \st(7.29) \ \\ 
            \bottomrule
        \end{tabular}
    }
    \vspace{-1.5\baselineskip}
\end{wrapfigure}

Preserving localized details is critical for brain tumor imaging. We therefore evaluated downstream tumor segmentation by feeding the synthesized T1ce images (T1, T2, and FLAIR $\rightarrow$ T1ce) into a pre-trained nnUNet \cite{isensee2021nnu}.
The quantitative results are summarized in Table \ref{table4}. Our method achieved the highest Dice scores for whole tumor (WT), enhancing tumor (ET), and tumor core (TC). Consistent with the baselines, ET and TC proved relatively more challenging to predict.
Qualitative results are visualized in Fig \ref{fig6}. Models such as MM-GAN, CCR, and MMHVAE failed to capture the ET and TC regions, producing overly dark tumor contrasts. ResViT generated distorted tumor shapes. In contrast, our model reconstructed tumor details more accurately, yielding segmentation predictions closest to the ground truth. Moreover, as shown in supplementary Fig. \ref{figa}, reducing $N_\text{EX}$ to 1 only slightly degrades downstream performance, indicating that the performance improvement is not solely due to expectation approximation sampling.

\begin{table} [t]
    \setlength{\tabcolsep}{4pt} 
    \centering
    \caption{Ablation study on training strategies and sampling methods. MTM and MTO denote many-to-many and many-to-one sampling, respectively. }
    \label{table5}
    \centering
    \scalebox{0.6}{
        \begin{tabular}{ccc|cccc|cccc}
            \toprule
            \multicolumn{3}{c|}{\textbf{Task}} & \multicolumn{4}{c|}{$\text{T}_{1} \rightarrow \text{T}_{2}$} & \multicolumn{4}{c}{$\text{T}_{1}, \text{T}_2 \rightarrow \text{T}_{2f}$} \\
            \cmidrule(lr){1-3} \cmidrule(lr){4-7} \cmidrule(lr){8-11} 
            \textbf{Training} & \textbf{MTM} & \textbf{MTO} & \textbf{PSNR} $\uparrow$ & \textbf{SSIM} $\uparrow$ & \textbf{LPIPS} $\downarrow$ & \textbf{FID} $\downarrow$ & \textbf{PSNR} $\uparrow$ & \textbf{SSIM} $\uparrow$ & \textbf{LPIPS} $\downarrow$ & \textbf{FID} $\downarrow$  \\
            \midrule
            \multirow{2}{*}{Full-modal only} & \checkmark & & 22.28 \st(2.49) & 85.95 \st(3.66) & 0.118 \st(0.028) & 83.87 & 23.76 \st(2.50) & 87.66 \st(3.18) & 0.129 \st(0.024) & 107.09 \\
                                             & & \checkmark & 22.19 \st(2.63) & 82.49 \st(5.81) & 0.225 \st(0.027) & 81.64 & 23.71 \st(2.34) & 85.25 \st(5.11) & 0.141 \st(0.023) & 110.10 \\
            \midrule
            \multirow{2}{*}{All scenarios}   & \checkmark & & 22.26 \st(2.15) & 83.69 \st(3.07) & 0.123 \st(0.022) & 65.75 & 23.86 \st(2.29) & 87.95 \st(3.02) & 0.122 \st(0.024) & 95.67  \\
                                             & & \checkmark & \blue{23.56 \st(2.70)} & \blue{88.47 \st(3.89)} & \blue{0.101 \st(0.029)} & \blue{54.15} & \blue{24.51 \st(2.33)} & \blue{88.46 \st(3.24)} & \blue{0.109 \st(0.026)} & \blue{87.96}  \\
            \bottomrule
        \end{tabular}
    }
\end{table}

\begin{figure}[t]
    \vspace{3pt}
    \centering
   \includegraphics[width=0.85\textwidth]{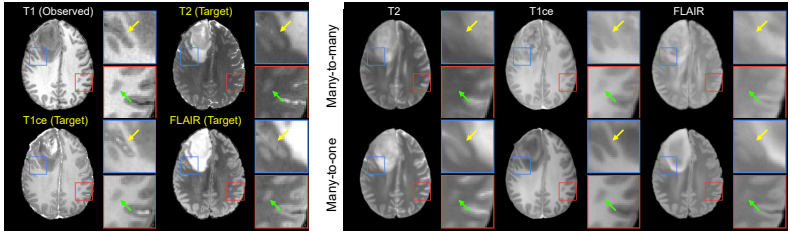}
    \caption{Visual comparison between many-to-many and many-to-one sampling given an observed T1. The blue and red bounding boxes highlight different anatomical regions, while the yellow and green arrows indicate distinct cortex areas.}
    \label{fig7}
\end{figure}

\subsection{Ablation Study}
We conducted an ablation study on our training and sampling strategies, with quantitative results summarized in Table \ref{table5}. For training, we compared a full-modal only (training on four modalities simultaneously from noise) against an all-scenarios (training on all combinations of two, three, or four modalities). For sampling, we evaluated many-to-many (MTM) sampling, which generates all missing modalities at once from the observed ones, versus many-to-one (MTO) sampling, which generates a single target modality. When trained on the full-modal only setup, the performance difference between MTM and MTO was marginal. This is likely because MTO requires masking other targets during sampling—a condition the model never encountered during full-modal training. However, when trained on All scenarios, MTO outperformed MTM in both image fidelity and quality. Visual comparisons between MTM and MTO are provided in Fig. \ref{fig7}. When synthesizing T2, T1ce, and FLAIR from a single T1 image, MTM failed to capture fine cortical details, whereas MTO preserved these subtle signals. These findings suggest that the iterative generation process in MTM is vulnerable to cross-modality interference, causing signal fading or hallucinations. The supplementary Fig.~\ref{figc} shows this error propagation across sampling process. 

\subsection{Perception-Distortion Trade-off}
\begin{wrapfigure}{r}{0.35\textwidth}
    \vspace{-1.5\baselineskip}
    \centering
    \includegraphics[width=0.35\columnwidth]{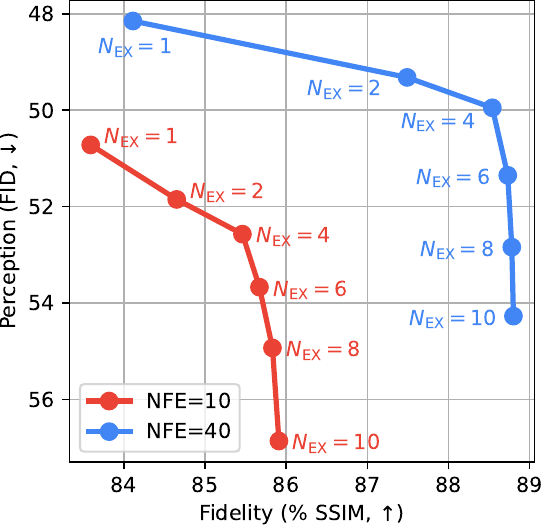}
    \caption{The perception-distortion trade-off curve.}
    \vspace{-1.5\baselineskip}
    \label{fig8}
\end{wrapfigure}

Averaging multiple samples via expectation approximation sampling reduces noise and enhances the main signal. However, because the ground truth inherently contains textured noise, increasing $N_\text{EX}$ improves image fidelity (lower distortion) at the cost of perceptual image quality. We observed the perception-distortion trade-off \cite{blau2018perception} during T1 $\rightarrow$ T2 synthesis (Fig. \ref{fig8}). At both NFE=10 and NFE=40, the highest perceptual quality is achieved at $N_\text{EX}=1$. As $N_\text{EX}$ increases, image fidelity increases while perceptual quality degrades. Furthermore, both quality and fidelity were notably higher at NFE=40 compared to NFE=10. This overall improvement is attributed to the measurement-guided vector field replacement (Eq. \ref{eq5}). Based on these findings, we empirically set NFE to 40 for optimal efficiency. Additional qualitative results across various NFE settings are provided in the supplementary material.

\section{Conclusion}
In this study, we demonstrate that complex architectures are unnecessary to build a generator capable of handling all missing modality scenarios. By simply training a flow-based model that iteratively samples data from noise, we showed that any missing modality can be effectively generated via posterior sampling.
Furthermore, our approach achieved superior image fidelity compared to existing baselines. It also yielded the highest performance in downstream tumor segmentation, proving its capability to accurately synthesize not only global anatomical structures but also critical tumor regions with high fidelity to the ground truth.

\section*{Acknowledgement} 
We sincerely thank Professor \textbf{Hyunjin Park} for his valuable guidance, advice, and support throughout this research.
This study was supported by the SMC-SKKU Future Convergence Research Program Grant (SMO125123). It was also supported by the AI Graduate School Support Program (Sungkyunkwan University) (RS-2019-II190421), the National Research Foundation of Korea (RS-2024-00408040), the ICT Creative Consilience Program Grant (IITP-2026-RS-2020-II201821), and the Institute of Information \& Communications Technology Planning \& Evaluation(IITP) grant (RS-2026-25528384, Resource-Intensive AI Technologies Based on Sustainable GPU Integrated Platforms), all funded by the Ministry of Science and ICT (MSIT), Republic of Korea.

% ---- Bibliography ----
%
% BibTeX users should specify bibliography style 'splncs04'.
% References will then be sorted and formatted in the correct style.
%
\bibliographystyle{splncs04}
\bibliography{main}

% supplementary
\clearpage

\title{Posterior Samplings are Missing Modalities Generators for Medical Image Translation \texorpdfstring{\\}{ } – Supplementary Material – } 
\author{}
\institute{}
\maketitle

\renewcommand{\thesection}{\Alph{section}}
\renewcommand{\thefigure}{\Alph{figure}}
\renewcommand{\thetable}{\Alph{table}}

\setcounter{section}{0}
\setcounter{figure}{0}
\setcounter{table}{0}
\setcounter{equation}{7}

% \section*{Overview}

\section{Formulations of Posterior Sampling Methods}

In inverse-problem formulations, the sampling trajectory is guided by a measurement consistency objective, as shown in Eq.~\ref{eq6}. The implementation details for each method are provided below.
For PSLD, FlowChef, and FlowDPS, we explicitly adapt their latent components for compatibility with our framework. In prior studies, these methods are implemented with latent space. We replace latent space guidance with pixel space guidance so that all methods operate consistently on the pixel space.

\vspace{3pt}
\noindent \textbf{Diffusion Posterior Sampling (DPS) \cite{chung2023diffusion}.} The update rule follows the original formulation in Eq.~\ref{eq6}. The step size $s$ is fixed to 1.

\vspace{3pt}
\noindent \textbf{Denoising Diffusion Null-space Model (DDNM) \cite{wang2023zeroshot}.} DDNM replaces Eq.~\ref{eq6} with a projection onto the null space of the measurement operator. The update formula is:
\begin{equation}
    \hat{\boldsymbol{x}}_{1} \leftarrow \mathcal{A}^\dagger\boldsymbol{y} + (\boldsymbol{I} - \mathcal{A}^\dagger\mathcal{A})\hat{\boldsymbol{x}}_{1},    \label{eq8}
\end{equation}
where $\mathcal{A}$ is the measurement operator and $\mathcal{A}^\dagger$ is its pseudo-inverse. We replace line 7 (Eq. \ref{eq6}) of Algorithm~\ref{algo1} with this update (Eq. \ref{eq8}). This projection-based correction enforces strict data consistency without requiring an explicit gradient term or the step size $s$.

\vspace{3pt}
\noindent \textbf{Posterior Sampling with Latent Diffusion (PSLD) \cite{rout2023solving}.} PSLD is originally designed for latent diffusion, with encoder-decoder mappings between pixel and latent space. We revise this latent formulation to a pixel space for implementation. We remove encoder-decoder components and compute guidance directly on image-space estimates. The residual loss $\mathcal{L}_{res}$ is defined as a weighted combination of the measurement residual and a projection residual:
\begin{equation}
\mathcal{L}_{res} = \| \boldsymbol{y} - \mathcal{A}\hat{\boldsymbol{x}}_{1}\|^2_2 + \gamma \|\mathcal{A}^T\boldsymbol{y} - \mathcal{A}^T\mathcal{A}\hat{\boldsymbol{x}}_{1}\|^2_2.
\end{equation}
The estimate is then updated using the gradient of this combined residual:
\begin{equation}
    \boldsymbol{\hat{x}}_1 \leftarrow {\boldsymbol{\hat{x}}_1} - s\nabla_{\boldsymbol{\hat{x}}_1} \mathcal{L}_{res},
    \label{eq10}
\end{equation}
where the residual weight is set to $\gamma=0.1$ and the step size is set to $s=1$. We replace line 7 (Eq. \ref{eq6}) of Algorithm~\ref{algo1} with this (Eq. \ref{eq10}) update.

\vspace{3pt}
\noindent \textbf{FlowChef \cite{Patel_2025_ICCV}.} We revise the latent implementation to a pixel space formulation. The guidance step updates $\hat{\boldsymbol{x}}_1$ by minimizing measurement error:
\begin{equation}
\hat{\boldsymbol{x}}_1 \leftarrow {\boldsymbol{\hat{x}}_1} - s\nabla_{\hat{\boldsymbol{x}}_1}\mathcal{L}(\hat{\boldsymbol{x}}_1, \boldsymbol{y}), \quad \mathcal{L}(\hat{\boldsymbol{x}}_1, \boldsymbol{y}) = \|\mathcal{A}\hat{\boldsymbol{x}}_{1} - \boldsymbol{y}\|^2_2
\label{eq11}
\end{equation}
We set the step size $s$ to 30 by grid search. We replace line 7 (Eq. \ref{eq6}) of Algorithm~\ref{algo1} with this (Eq. \ref{eq11}) equation. 

\vspace{3pt}
\noindent \textbf{Flow-Driven Posterior Sampling (FlowDPS) \cite{Kim_2025_ICCV}.} For FlowDPS, we similarly revise latent components to a pixel space. FlowDPS introduces an inner optimization loop at each flow step. Instead of applying a single gradient update, it performs multiple gradient descent iterations directly on $\hat{\boldsymbol{x}}_{1|t}$. For $k = 1, \dots, K$:
\begin{equation}
\hat{\boldsymbol{x}}_{1|t}^{(k)} \leftarrow \hat{\boldsymbol{x}}_{1|t}^{(k-1)} - s \nabla_{\hat{\boldsymbol{x}}_{1}} \|\boldsymbol{y} - \mathcal{A}(\hat{\boldsymbol{x}}_{1|t}^{(k-1)})\|^2_2, \quad \hat{\boldsymbol{x}}_1 \leftarrow \hat{\boldsymbol{x}}_{1|t}^{(K)}.
\label{eq12}
\end{equation}
We set the number of gradient steps $K$ to 3 and step size $s$ to 30 by grid search. We replace line 7 (Eq. \ref{eq6}) of Algorithm~\ref{algo1} with this (Eq. \ref{eq12}) update.

\section{Additional Results}

\begin{figure}[h]
    \vspace{-12pt}
    \centering
    \includegraphics[width=0.99\textwidth]{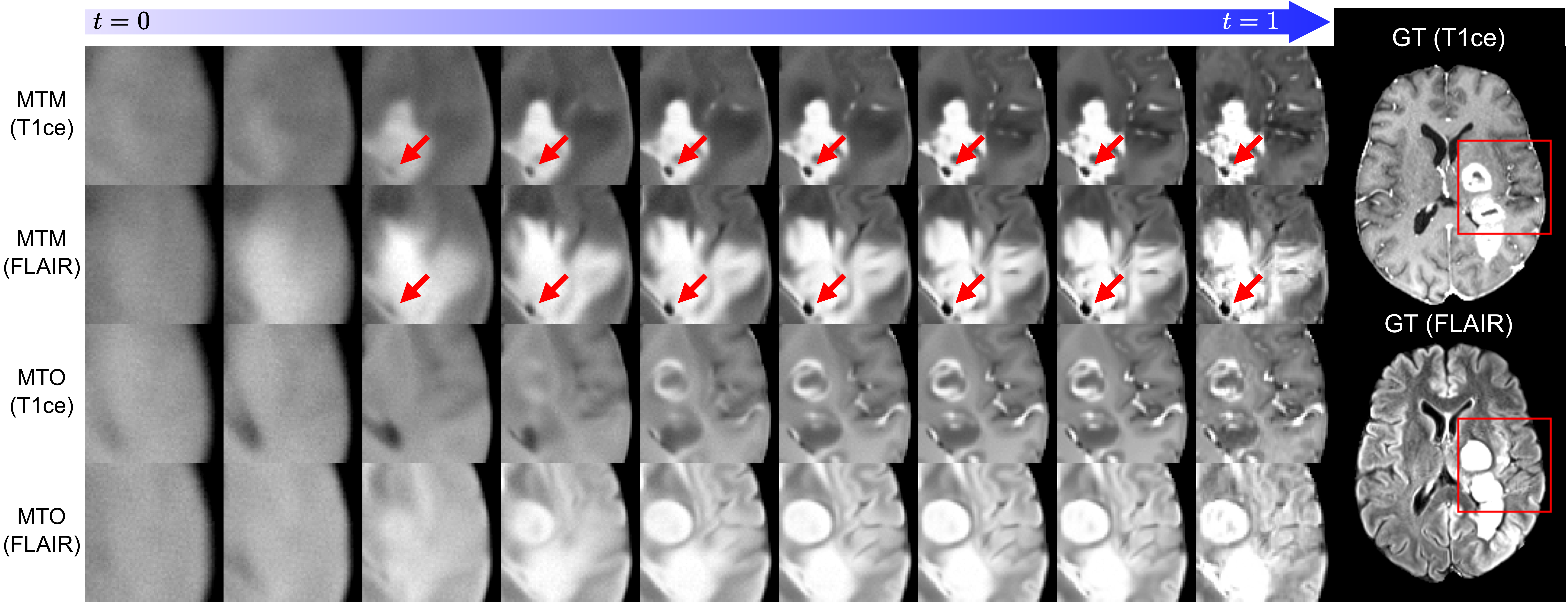}
    \vspace{-6pt}
    \caption{Visualization of error propagation in MTM and MTO ($\text{T}_{1}, \text{T}_2 \rightarrow \text{T}_{1ce}, \text{T}_{2f}$).}
    \label{figc}
    \vspace{-6pt}
\end{figure}

\noindent \textbf{Visualization of error propagation.} We visualized $\hat{\boldsymbol{x}}_1$ at various time steps $t$ for both MTM and MTO (Fig. \ref{figc}). While MTO generates images without apparent error propagation, the red arrows in MTM highlight hallucinations in which errors spread during simultaneous generation.

\begin{figure}[h]
    \vspace{-12pt}
    \centering
    \includegraphics[width=0.99\columnwidth]{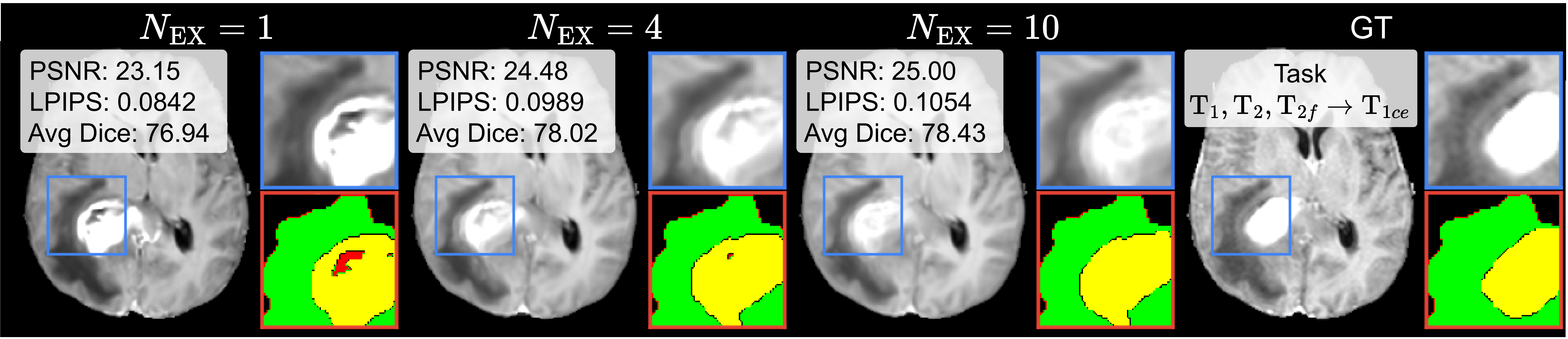}
    \vspace{-6pt}
    \caption{Evaluation on the downstream task with various $N_\text{EX}$.}
    \label{figa}
    \vspace{-12pt}
\end{figure}

\noindent \textbf{Additional Segmentation Results.} Even with $N_\text{EX}=1$, our method still outperforms the comparison models, with an average Dice of 76.94\%. However, performance improves further as $N_\text{EX}$ increases (Fig. \ref{figa}). 
An averaged image with reduced noise and clearer signal positively affects downstream-task performance.

\begin{figure}[t]
    \centering
    \includegraphics[width=0.99\textwidth]{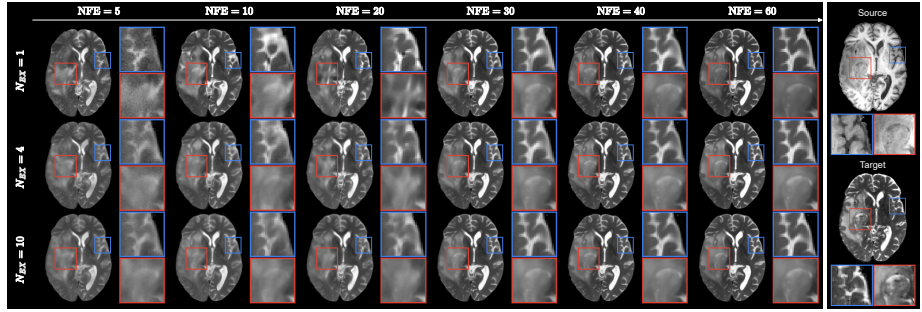}
    \vspace{-6pt}
    \caption{Visualization of the perception-distortion trade-off. Increasing NFE improves image quality, while increasing $N_\text{EX}$ enhances image fidelity. The red and blue bounding boxes indicate the cortex and tumor regions, respectively.}
    \label{figA}
    \vspace{-12pt}
\end{figure}

\noindent \textbf{Perception-Distortion Trade-off.}
Fig. \ref{figA} shows the results for different NFE and $N_\text{EX}$ in T1$\rightarrow$T2 synthesis. For NFE=5, the noise is prominent when $N_\text{EX}=1$ but progressively decreases with higher $N_\text{EX}$. We observe that higher NFE yields better image quality, while higher $N_\text{EX}$ leads to greater image fidelity to the target, highlighting the perception-distortion trade-off.

\begin{figure}[h]
    \vspace{-12pt}
    \centering
    \includegraphics[width=0.8\textwidth]{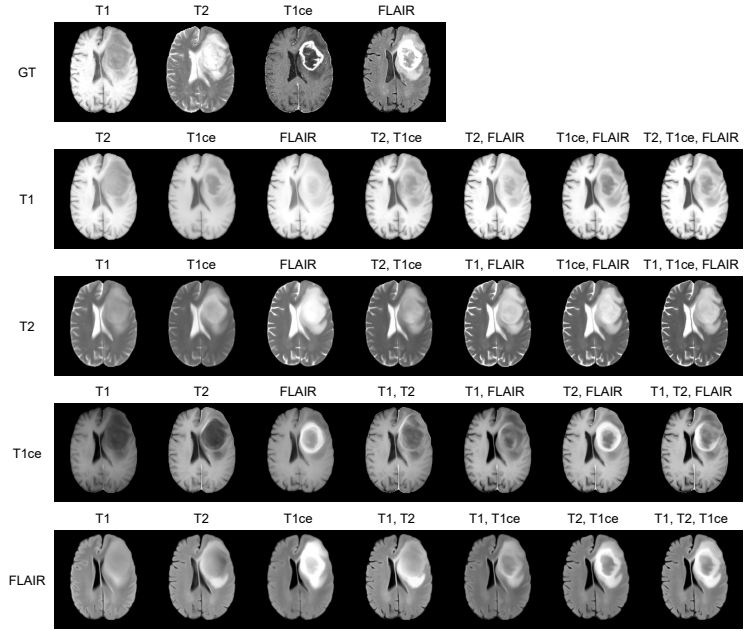}
    \vspace{-6pt}
    \caption{Qualitative results of synthesized targets in various scenarios. An increased number of sources leads to a more precise delineation of the tumor region, improved structural detail and image clarity. }
    \label{figB}
    \vspace{-12pt}
\end{figure}

\vspace{3pt}
\noindent \textbf{Additional Qualitative Results.} Fig. \ref{figB} shows the targets generated from different sources. This allows for a qualitative comparison across various source-target scenarios.

\end{document}